%% file: main.tex
\newif\ifcomments  %
\newif\ifsupp  %
\newif\ifresolvedcomments
\commentstrue
\supptrue
\resolvedcommentsfalse

\documentclass{article}

\usepackage{microtype}
\usepackage{graphicx}
\usepackage{subfigure}
\usepackage{booktabs} %
\usepackage{amssymb}
\usepackage{amsmath}
\usepackage[ruled,vlined]{algorithm2e}
\usepackage{comment}
\usepackage{dblfloatfix}
\newcommand{\argmin}{\operatornamewithlimits{argmin}}

\usepackage{hyperref}
\usepackage{xcolor}
\usepackage{tablefootnote}

\newcommand\numberthis{\addtocounter{equation}{1}\tag{\theequation}}
\newtheorem{lem}{Lemma}[section]
\newtheorem{definition}[lem]{Definition}

\def\longname{Hessian-Free Gradients Matching}
\def\shortname{HFGM}

\usepackage[accepted]{icml2021}

\icmltitlerunning{A Method to Reveal Speaker Identity in Distributed ASR Training, and How to Counter It}

\begin{document}

\twocolumn[
\icmltitle{A Method to Reveal Speaker Identity in Distributed ASR Training, \\ and How to Counter It}

\icmlsetsymbol{equal}{*}

\begin{icmlauthorlist}
\icmlauthor{Trung Dang}{goo2,bos}
\icmlauthor{Om Thakkar}{goo}
\icmlauthor{Swaroop Ramaswamy}{goo}
\icmlauthor{Rajiv Mathews}{goo}
\icmlauthor{Peter Chin}{bos}
\icmlauthor{Françoise Beaufays}{goo}
\end{icmlauthorlist}

\icmlaffiliation{goo}{Google}
\icmlaffiliation{bos}{Boston University}
\icmlaffiliation{goo2}{Work performed while at Google.}

\icmlcorrespondingauthor{Trung Dang}{trungvd@bu.edu}

\icmlkeywords{gradients matching, ASR, distributed training}
\vskip 0.3in]

\printAffiliationsAndNotice{}  %

\input{sections/abstract}

\input{sections/intro}

\input{sections/background}

\input{sections/proposed}

\input{sections/experiments}

\input{sections/defense}

\input{sections/multi}

\input{sections/prior}

\input{sections/acks}

\bibliographystyle{icml2021}
\bibliography{strings}

\ifsupp
\appendix
\input{sections/appendix/background}

\input{sections/appendix/expt}
\input{sections/appendix/app_proposed}
\input{sections/appendix/visualize}
\fi 

\end{document}

%% file: sections/abstract.tex
\begin{abstract}

End-to-end Automatic Speech Recognition (ASR) models are commonly trained over spoken utterances using optimization methods like Stochastic Gradient Descent (SGD). 
In distributed settings like Federated Learning, model training requires transmission of gradients over a network. In this work, we design the first method for revealing the identity of the speaker of a training utterance with access only to a gradient.
We propose \longname{}, an input reconstruction technique that operates without second derivatives of the loss function (required in prior works), which can be expensive to compute. 
We show the effectiveness of our method using the DeepSpeech model architecture, demonstrating that it is possible to reveal the speaker’s identity with 34\% top-1 accuracy (51\% top-5 accuracy) on the LibriSpeech dataset.
Further, we study the effect of two well-known techniques, Differentially Private SGD and Dropout, on the success of our method.
We show that a dropout rate of 0.2 can reduce the speaker identity accuracy to 0\% top-1 (0.5\% top-5).
\end{abstract}

%% file: sections/intro.tex
\section{Introduction}
\label{sec:intro}

End-to-end automatic speech recognition (ASR), which directly transcribes speech to text without predefined alignments, has increasingly been gaining popularity over conventional pipeline frameworks. 
State-of-the-art ASR models have achieved human parity in conversational speech recognition \cite{xiong2016achieving}.
Training such models often requires a large amount of user-spoken utterances. 
In the speech domain, training data includes audio and transcripts of utterances, which can directly expose sensitive information, or make it possible to leak attributes such as gender, dialect, or identity of the speaker. 

In distributed frameworks such as Federated Learning (FL) \citep{mcmahan2017federated}, model training is performed via mobile devices with transmission of gradients over the network, allowing training over large populations \cite{bonawitz2019towards} while ensuring such data remains on-device. 
 Many works have shown the competitive performance of FL-trained models on sequential modeling tasks like keyboard prediction \cite{hard2018federated} and keyword spotting \cite{leroy2019federated,hard2020training}, as well as in speech recognition~\cite{dimitriadis2020federated,guliani2020training}. 
 
A recent line of work \cite{zhu2019deep,geiping2020inverting,wei2020framework} has focused on demonstrating leakages of  information about training data, from the gradients used in model training.
At a high level, these works aim to reconstruct training samples by designing optimization methods for constructing objects that have a  gradient \emph{matching} to the observed gradient.
For instance, a number of existing methods have been shown to successfully reconstruct images used for training image classification models.
As we discuss later (in Section~\ref{sec:comp}), there are  fundamental challenges, like variable-sized inputs/outputs, that render such methods inapplicable in the speech domain.

In this work, we study information leakage from gradients in ASR model training. 
In particular, we design a method to reveal the speaker identity of a training utterance from a model gradient computed using the utterance.
Given that ASR models can have training utterances and transcripts of arbitrary lengths, for computational efficiency and to avoid potential false positives, we assume that the transcript, and the length of the training utterance are known. 
We start by designing \longname{} (\shortname{}), a technique to reconstruct speech features used in computing gradients for training ASR models.
Our \shortname{} technique eliminates the need of second-order derivatives (i.e., Hessian) of the loss function, which were required in prior works~\cite{zhu2019deep,geiping2020inverting,wei2020framework}, and can be expensive to compute. 
Next, our method uses the reconstructed features and a speaker identification model to uniquely identify the speaker from a list of speakers by comparing speaker embeddings.

To our knowledge, this is the first method in the speech domain that can be used for revealing information about training samples from gradients. 
We demonstrate the efficacy of our method by conducting experiments using the LibriSpeech data set \cite{panayotov2015librispeech} on the DeepSpeech \cite{hannun2014deep} model architecture.
We find that our method is successful in revealing the speaker identity with 34\% top-1 accuracy (51\% top-5 accuracy) among $\approx$2.5k speakers.

We also study the effect of two standard training techniques, namely, Differentially Private Stochastic Gradient Descent (DP-SGD)~\cite{BST14, abadi2016deep}, and Dropout \cite{srivastava2014dropout}, on the success of our method.
The technique of DP-SGD is the state-of-the-art in training deep neural networks with Differential Privacy (DP)~\cite{DMNS, ODO} guarantees.
Intuitively, DP prevents an adversary from confidently making any conclusions about whether any particular sample was used to train a model, even while having access to the model and arbitrary external side information.
While we demonstrate that using DP-SGD can mitigate the success of our method, we find (in line with prior works~\cite{abadi2016deep, mcmahan2017learning, thakkar2020understanding}) that training large models using DP-SGD can significantly affect model utility.

The well-known technique of Dropout \cite{hinton2012improving,srivastava2014dropout}, which randomly drops hidden units of a neural network during training, is commonly employed to avoid overfitting in large models. We show that using dropout can reduce the speaker identity accuracy of our method to 0\% top-1 (0.5\% top-5), without compromising utility of the trained model. 

We make the following contributions:
\begin{enumerate}
    \item We design the first method to reveal speaker identity of an utterance in ASR model training, with access to only a gradient computed using the utterance. We achieve this via  \longname{}, an input reconstruction technique that operates without needing second derivatives of the loss function.
    \item We empirically demonstrate the effectiveness of our method, using the DeepSpeech model architecture, in revealing speaker identity with 34\% top-1 accuracy (51\% top-5 accuracy) on the LibriSpeech data set. 
    To spur further research, we provide an open-source implementation\footnote{\url{https://github.com/googleinterns/deepspeech-reconstruction}} of our experimental framework.
    \item We study the effect of two standard training techniques -- DP-SGD and Dropout -- on the success of our method.
    We empirically demonstrate that using dropout can reduce the success of our method to 0\% top-1  (0.5\% top-5) accuracy.
\end{enumerate}

We conclude by exploring the effectiveness of our method in two complex regimes, where instead of access to individual gradients, the method can access 1) only the average gradient from a mini-batch of samples, and 2) the update comprising multiple gradient descent steps using a training sample. 
We demonstrate that in both of the above settings, our method reveals speaker identity with non-trivial accuracy, whereas training with dropout is effective in reducing its success.

%% file: sections/background.tex
\section{Background}
\label{sec:background}

In this section, we start by revisiting the idea of Gradients Matching (GM) \cite{zhu2019deep}, which has been successfully applied to reconstruct images from gradients of an image recognition model. 
We then introduce end-to-end ASR models and a popular architecture using the Connectionist Temporal Classification (CTC) loss \cite{graves2006connectionist}. 
We also provide some background on zeroth-order optimization, specifically the direct search algorithm, and speaker identification models with triplet loss, which form critical components of our proposed method. 

\paragraph*{Gradients Matching \& Deep Leakage from Gradients (DLG) Algorithm} DLG was introduced by \citet{zhu2019deep} as a method to reconstruct an 
input $\hat{x}$ and output $\hat{y}$ given a model gradient $\nabla\mathcal{L}_\theta(\hat{x},\hat{y})$, where $\mathcal{L}$ denotes the loss function and $\theta$ denotes the model parameters (when it is clear from the context that the gradient is w.r.t. model parameters $\theta$, we just denote it by $\nabla\mathcal{L}(\cdot,\cdot)$).
The algorithm attempts to find an input-output pair $(x,y)$ that matches $\nabla\mathcal{L}(x,y)$ with $\nabla\mathcal{L}(\hat{x},\hat{y})$. The general idea is also referred to as Gradients Matching (GM).
A dummy input $x$ and a dummy label $y$ are fed into the model to get dummy gradients $\nabla\mathcal{L}(x,y)$. Reconstructed objects are obtained by minimizing the Euclidean distance between the dummy gradients and the client update.

\begin{equation}
x^*, y^*=\argmin_{x,y}\| \nabla \mathcal{L}(x,y) -\nabla \mathcal{L}(\hat{x},\hat{y}) \|^2\label{eqn:gm1}
\end{equation}

\citet{geiping2020inverting} provide an extension of DLG that works with larger images and trained models. They adopt the cosine similarity (also shown in \eqref{eqn:prob}) to optimize the gradients distance, which only matches the direction of gradients. A regularization term is also added.
They assume that the training labels are known and formulate the reconstruction as finding $x$ to minimize

\begin{equation}
\mathcal{D}(x, \nabla\mathcal{L}(\hat{x},\hat{y}))=1-\frac{\left<\nabla\mathcal{L}(x,\hat{y}),\nabla\mathcal{L}(\hat{x},\hat{y})\right>}{\|\nabla\mathcal{L}(x,\hat{y})\|\|\nabla\mathcal{L}(\hat{x},\hat{y})\|}+\alpha TV(x)\label{eqn:gm2}
\end{equation}
where $TV(x)$ and $\alpha$ are a regularization term (total variance) and its weight in the loss, respectively.

Both first-order and second-order optimization methods can be used to optimize \eqref{eqn:gm1} and \eqref{eqn:gm2}. While first order techniques, such as Stochastic Gradient Descent (SGD) and Adam optimizer are usually faster to compute, second-order methods like L-BFGS can escape slow convergence paths better \cite{battiti1992first}. An observation about this function is the presence of first-order derivatives $\nabla\mathcal{L}(x,\hat{y})$, which requires second-order differentiation with regards to $x$ to compute the gradients. These second-order gradients can usually be derived with auto-differentiation commonly implemented in deep learning frameworks.

\paragraph*{End-to-end ASR with CTC loss}
 
Recently, end-to-end models have achieved superior performance while being simpler than traditional pipeline models \cite{graves2006connectionist,graves2012sequence,chorowski2014end,chorowski2015attention,chan2015listen,xiong2016achieving}. Two major lines of end-to-end ASR architectures are based on connectionist temporal classification (CTC) \cite{graves2006connectionist}, and the attention-based encoder-decoder mechanism \cite{chorowski2014end}. 

We focus on models with CTC loss in this work. CTC models are able to align network inputs of length $T$ with label sequences of length $S \le T$ without predefined alignments. The set of labels $L$ is extended with a blank label, denoted as ``$\varnothing$", to form $L_\varnothing=L\cup\{\varnothing\}$. The label sequence $z$ can be mapped with a one-to-many mapping $\mathcal{P}: L^T\rightarrow L_\varnothing^S$ to CTC paths \cite{graves2006connectionist}, for example, ``\texttt{aab}'' can both be mapped to ``\texttt{a$\varnothing$$\varnothing$$\varnothing$ab$\varnothing$}'' or ``\texttt{$\varnothing$aa$\varnothing$abb}''

A neural network $\mathcal{F}$, usually including one or several bi-directional RNNs to model frame dependencies, is used to map a $d$-dimensional speech features $x\in\mathbb{R}^{T\times d}$ to $o=\mathcal{F}(x)\in\mathbb{R}^{T\times|L_\varnothing|}$. The probability of a CTC path $\pi\in \mathcal{P}(l)$ is defined as $p({\pi}|{x})=\prod_{t=1}^T o_{t,\pi_t}$. The likelihood of a label sequence $y$ follows as the sum over probabilities of all CTC paths:
\begin{equation}
    p(y|x)=\sum_{{\pi}\in \mathcal{P}(l)}p({\pi}|{x})\label{eqn:ctc}
\end{equation}
The model is optimized by maximizing the likelihood of ${y}$, i.e. minimizing $\mathcal{L}_{CTC}=-\sum_{({x},{y})}\ln p({y}|{x})$. During inference, a greedy search or beam search is conducted to find a sequence ${y}$ that maximizes $p({y}|{x})$

Note that the number of terms in \eqref{eqn:ctc} grows exponentially with the length of inputs. To optimize the loss, its value and first-order derivatives are computed analytically with a dynamic programming algorithm, namely the CTC forward-backward algorithm \cite{graves2006connectionist}.

\paragraph{Zeroth-order Optimization} Zeroth-order optimization is the process of minimizing an objective, given access to the objective values at chosen inputs. 
The standard approach to zeroth-order optimization is to estimate the gradients \cite{FKTM05}.
However, gradient estimation suffers from high variance due to non-robust local minima or highly non-smooth objectives \cite{golovin2019gradientless}. 
A direct search algorithm (also known as pattern search, derivative-free search, or black-box search), which samples a vector $u$ and moves $x$ to $x+u$, performs well in several settings, such as reinforcement learning \cite{mania2018simple}. Additionally, a binary search on the step size, referred to as Gradientless Descent \cite{golovin2019gradientless} is proven to be fast for high-dimensional zeroth-order optimization.

\paragraph*{Speaker Identification with Triplet Loss}
Recent approaches \cite{snyder2018x,chung2018voxceleb2}
 formulate speaker identification as learning speaker discriminative embeddings for an utterance. The embeddings are extracted from variable-length acoustic segments via a deep neural network. The multi-class cross entropy loss \cite{snyder2017deep,snyder2018x} and the triplet loss \cite{zhang2017end,li2017deep,chung2018voxceleb2} are two common approaches to train the embeddings.
 In this work, we adopt the triplet loss, which operates on pairs of embeddings, trying to minimize the distance of embeddings from the same speaker, and maximize the distance with other negative samples.

%% file: sections/proposed.tex
\section{A Method to Reveal Speaker Identity}
\label{sec:method}

\begin{figure}[t]
\begin{center}
\centerline{\includegraphics[width=0.5\textwidth]{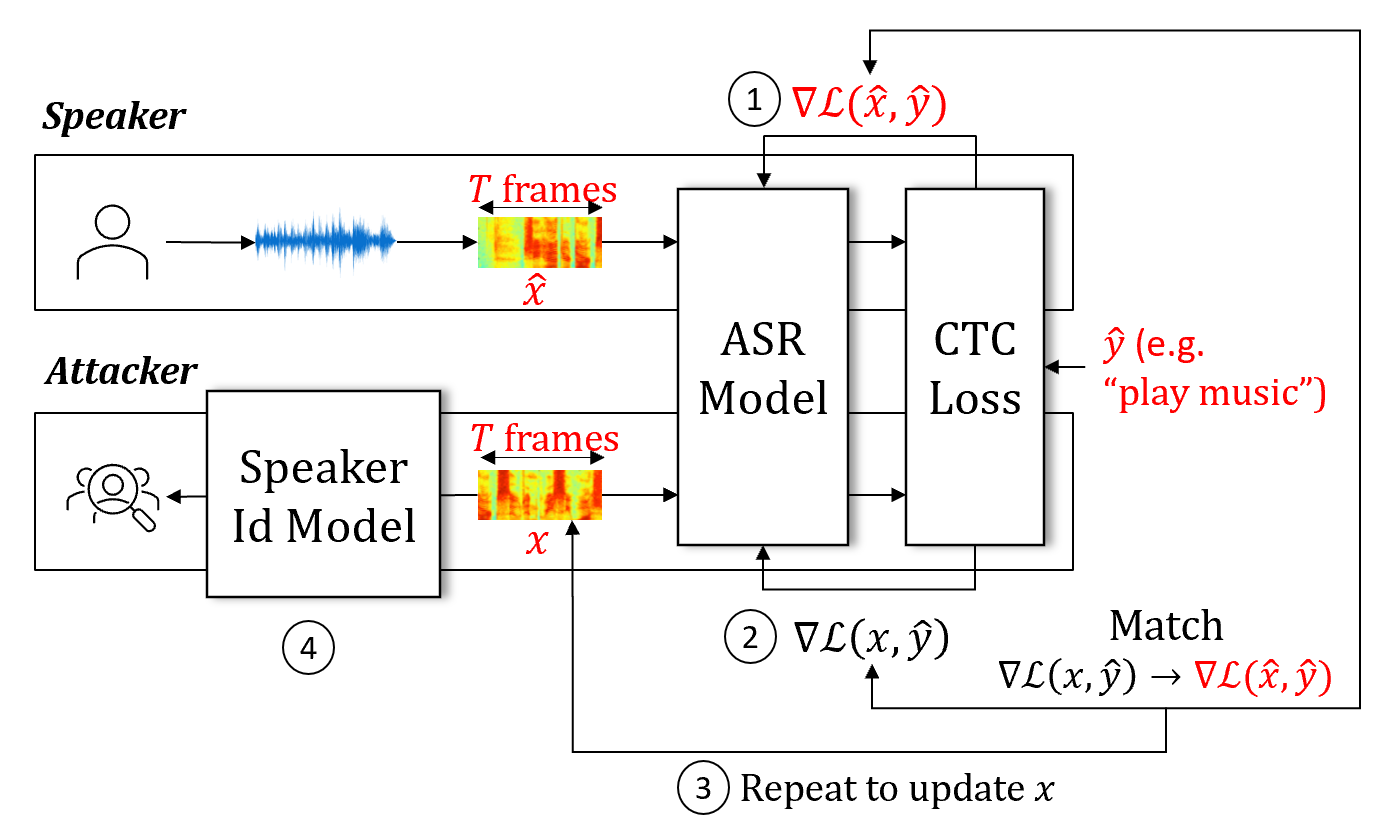}}
\caption{An illustration of our method. (1) A gradient is accessible to an attacker. (2) The attacker computes dummy gradients from a dummy input $x$. (3) The attacker compares the gradient received with dummy gradients and repeats to optimize $x$. (4) The attacker reveals the identity of the speaker. Notations in red are known to the attacker.}
\label{fig:attack}
\end{center}
\end{figure}

In this section, we describe our method to reveal the speaker identity of a training sample $\hat{x}$ given its model gradient $\nabla\mathcal{L}(\hat{x},\hat{y})$.
Here, $\hat{x} \in \mathbb{R}^{T\times d}$ denotes the input speech features created from the training utterance, $T$ is the length of input, $d$ is the dimension of the input speech features,  $\hat{y}$ is the output label sequence, and $\mathcal{L}$ denotes the training loss function.
We split our method into two phases: (1) Using \longname{} to reconstruct the input speech features (reconstruction phase), and (2) Identify the speaker from the reconstructed speech features (inference phase). 
Figure \ref{fig:attack} provides an illustration of our method.  

\subsection{Reconstruction Phase: \longname{} (\shortname{})}
\label{sec:dfgm}
Given access to a gradient $\nabla\mathcal{L}(\hat{x},\hat{y})$, generally one would like to find a pair of speech features and transcript $(x,y)$ such that $\nabla\mathcal{L}(x,y)=\nabla\mathcal{L}(\hat{x},\hat{y})$.
However, ASR models are typically sequence-to-sequence models that can map arbitrary length speech features ($\hat{x}$) to arbitrary length transcripts ($\hat{y}$). 
With no additional information, the possible values $y$ can take is exponential in the label set size, searching through which can incur a prohibitive computational cost.
Moreover, there can be many false positives, i.e., pairs $(x',y') \neq (\hat{x},\hat{y})$ such that $\nabla\mathcal{L}(x',y')=\nabla\mathcal{L}(\hat{x},\hat{y})$.
To circumvent this issue and make our problem simpler, we assume $\hat{y}$, the transcript of the training utterance,  is given.\footnote{Note that the transcript could be a common phrase, e.g., ``play music". Our objective is to identify the speaker of an utterance regardless of the contents of its transcript.}
Even though $\hat{y}$ is given, there could exist multiple length speech features $x'$, where $|x'| \neq T$, such that $\nabla\mathcal{L}(x',\hat{y})=\nabla\mathcal{L}(\hat{x},\hat{y})$.
Thus, we also assume $T$ is given.
Note that even if the transcript and the length of the input speech features are known, revealing the identity of the speaker can still result in a significant breach of privacy.
Designing efficient reconstruction methods that
operate without these assumptions is an interesting direction, which we leave for future work.\footnote{Using the experimental setup in Section~\ref{sec:results}, we provide some preliminary results in Appendix~\ref{app:assump} on reconstruction  i) without the knowledge of length of input speech features, and ii) with the knowledge of only the transcript length. We find that while reconstruction can succeed even with a good estimate of the input length, it fails with no knowledge of the transcript.}

Now, we define the reconstruction task as constructing an $x\in\mathbb{R}^{T\times d}$ such that $\nabla\mathcal{L}(x,\hat{y})$ is close to the observed gradient $\nabla\mathcal{L}(\hat{x},\hat{y})$.
Following \cite{geiping2020inverting}, we choose cosine distance as our measure of closeness, and formulate our optimization problem to find $x^*$ s.t.:

\begin{align*}
    x^*=\argmin_x\mathcal{D}\left(x,\nabla\mathcal{L}(\hat{x},\hat{y})\right), \qquad \text{where} \\
    \mathcal{D}(x,\nabla\mathcal{L}(\hat{x},\hat{y}))=1-\frac{\left<\nabla\mathcal{L}(x,\hat{y}), \nabla\mathcal{L}(\hat{x},\hat{y})\right>}{\|\nabla\mathcal{L}(x,\hat{y})\|\|\nabla\mathcal{L}(\hat{x},\hat{y})\|} \numberthis \label{eqn:prob}
\end{align*}

To solve this non-convex optimization problem using gradient-based methods as in prior work~\citet{zhu2019deep, geiping2020inverting, wei2020framework}, we need to compute the second-order derivatives of the loss function $\mathcal{L}$. 
The loss function that is of interest here is the CTC loss which is commonly used in end-to-end ASR systems.
However, computing the second derivative of CTC loss involves backpropagating twice through a dynamic programming algorithm which we found to be intractable.\footnote{Additionally, the second derivatives of CTC loss are not implemented in common deep learning frameworks like TensorFlow~\cite{tensorflow2015-whitepaper} and PyTorch~\cite{pytorch}.}
 To tackle this challenge, and also address a broader family of loss functions, we adopt a zeroth-order optimization algorithm.

\begin{algorithm}[h]
   \caption{\longname{}}
   \label{alg:dfgm}
\begin{algorithmic}
   \STATE {\bfseries Input:} Gradients to match $\nabla\mathcal{L}(\hat{x},\hat{y})$, gradients distance function $\mathcal{D}(x,\nabla\mathcal{L}(\hat{x},\hat{y}))$, learning rate $\alpha$, transcript $\hat{y}$ and length of speech features $T$. Parameters: number of samplings $k$, number of iterations $N$
   \STATE Initialize $x\in\mathbb{R}^{T\times d}$.
   \FOR{$n=1$ {\bfseries to $N$}}
   \STATE $V\leftarrow\varnothing$
   \STATE Sample $k$ unit vectors $v_1,...,v_k\in\mathbb{R}^{T\times d}$
   \FOR{$k=1$ {\bfseries to} $K$}
        \IF{$\mathcal{D}(x+\alpha v_k,\nabla\mathcal{L}(\hat{x},\hat{y}))<\mathcal{D}(x,\nabla\mathcal{L}(\hat{x},\hat{y}))$}
            \STATE Add $v_k$ to $V$
        \ENDIF
   \ENDFOR
   \STATE $x\leftarrow x+\alpha \sum_{v\in V}v$
   \ENDFOR
\end{algorithmic}
\end{algorithm}

We use a direct search approach (Section \ref{sec:background}) called \shortname{} (Algorithm~\ref{alg:dfgm}). We initialize $x$ with uniformly random values. At each iteration, we sample $k$ random unit vectors and apply them to $x$. The value $\mathcal{D}$ is evaluated at each of these $k$ points. We choose only the vectors that lower $\mathcal{D}$, sum these up and apply the sum with a learning rate $\alpha$. We repeat the process until we reach the convergence criteria.

\subsection{Inference Phase: Revealing Speaker Identity}

In the second part of our method, we use the reconstructed speech features ($x^*$) to identify the speaker of the utterance from a list of possible speakers.

We train a speaker identification model that uses the same speech features as our ASR model on some corpus. We assume that we have access to some public utterances for each possible speaker to identify them. We use the speaker identification model to create embeddings for each  speaker from the public utterances. We take the reconstructed speech features ($x^*$), create an embedding using the speaker identification model, and compare it with embeddings for each speaker. 
If the method is successful, the embedding created from $x^*$ is closer to the embedding for the speaker of the utterance than the other speakers.

\subsection{Comparison with Related Prior Works}
\label{sec:comp}

Our work differs from the related prior works~ \cite{zhu2019deep,geiping2020inverting,wei2020framework} in a few ways.

\textbf{Input features:} The input to ASR models is typically not the raw audio but speech features which are computed from the raw audio using a series of lossy transformations. In image recognition models, the input to the model is typically the raw pixel values. While prior works on image recognition models demonstrate breach of privacy by directly reconstructing the input to the model, we incorporate an additional inference phase where we use the reconstructed input to reveal the identity of the speaker.

\textbf{Variable-sized inputs and outputs:} We focus on ASR models, which have variable-sized inputs and outputs; image recognition models have fixed size inputs and outputs. 

\textbf{CTC Loss:} The models we focus on use CTC loss instead of cross-entropy loss. CTC loss is significantly more complex, requiring a dynamic programming algorithm to compute the value of the loss function and derivatives.

%% file: sections/experiments.tex
\section{Experiments}
\label{sec:results}

In this section, we provide empirical results on the effectiveness of our method.
We start by describing our setup.

\paragraph{Model Architectures} Following prior work~\cite{carlini2018audio}, we choose the DeepSpeech \cite{hannun2014deep} model architecture for our experiments.
The model consists of three feed-forward layers, followed by a single bi-directional LSTM layer, and two feed-forward layers to produce softmax probabilities for the CTC loss. 
DeepSpeech uses character-based CTC loss: the output is a sequence of characters. The input is a 26-dimensional mel-frequency cepstrum coefficients (MFCCs) feature. 
MFCCs are a popular speech feature, derived by mapping the power of the result of a Fourier transform to the mel scale, then performing a discrete cosine transformation. We use Mozilla's implementation of DeepSpeech\footnote{\url{https://github.com/mozilla/DeepSpeech}}

We conduct our experiments using randomly initialized weights for the model.
For the inference phase, we follow \citet{li2017deep} to train a text-independent speaker identification model on 26-dim normalized MFCCs, similar to the speech feature used for inputs to DeepSpeech. 

\paragraph{Dataset}We choose the LibriSpeech ASR corpus \cite{panayotov2015librispeech}, a large-scale benchmark speech dataset for our experiments. The dataset contains pairs of audio and transcript, along with speaker attributes such as gender and identity.
For training the speaker identification model, we first combine all the dev-\{clean/other\}, test-\{clean/other\}, train-\{clean-100/clean-360/other-500\} sets to obtain 300k utterances from 2,484 speakers, and use the first 5 utterances of each speaker for training. 

For the reconstruction phase, we trim the leading and ending silences, based on the intensity of every 10ms chunk, from each utterance in the remaining combined test set. Next, we randomly sample a total of 600 utterances, 100 for each interval of audio length in $\{[1,1.5s), [1.5,2s), \ldots, [3.5,4s)\}$.
The average audio length in our sampled set is 2.5 seconds, and average transcript length is 40.6 characters. 
The male:female ratio in the sampled utterances is 1.1:1.

\paragraph{Implementation Details}
In this section, all the experiments consider the scenario of revealing speaker identity from a single gradient computed using a single utterance.
For computational efficiency, we match gradients only for the last layer ($\sim$60k parameters). 
Note that matching lower layers may increase the reconstruction quality.

Each dummy input in our reconstruction is initialized with uniformly random values in $[-1, 1]$. 
When performing direct search, we sample 128 unit vectors per iteration, each of which only updates a single frame. We set the step size to 1, and reduce by half after every 2.5k iterations s.t. the loss does not decrease by more than 5\%. 
We stop the reconstruction when the step size reaches 0.125. We run each reconstruction on a single Tesla V100 GPU. The reconstruction time depends on the length of inputs, ranging from 3 to 6 hours. %

\paragraph{Evaluation Metrics} 
To evaluate our reconstruction, we use the Mean Absolute Error (MAE) to measure the distance of normalized MFCCs to those of the original utterance. 
During inference, the similarity scores of a reconstructed object's embedding with each of 5 available utterances' embeddings in the training data are averaged and ranked to identify the speaker.
We use Top-1 Accuracy, Top-5 Accuracy and Mean Reciprocal Rank (MRR) to evaluate the speaker identity leakage. In experiments where we try to use alternate training methods, the Word Error Rate (WER) is used to evaluate the quality of trained ASR models\footnote{An N-gram language model is trained separately on a large text corpus and used during inference.}.

\begin{figure*}[htb]
\begin{center}
\centerline{\includegraphics[width=\textwidth]{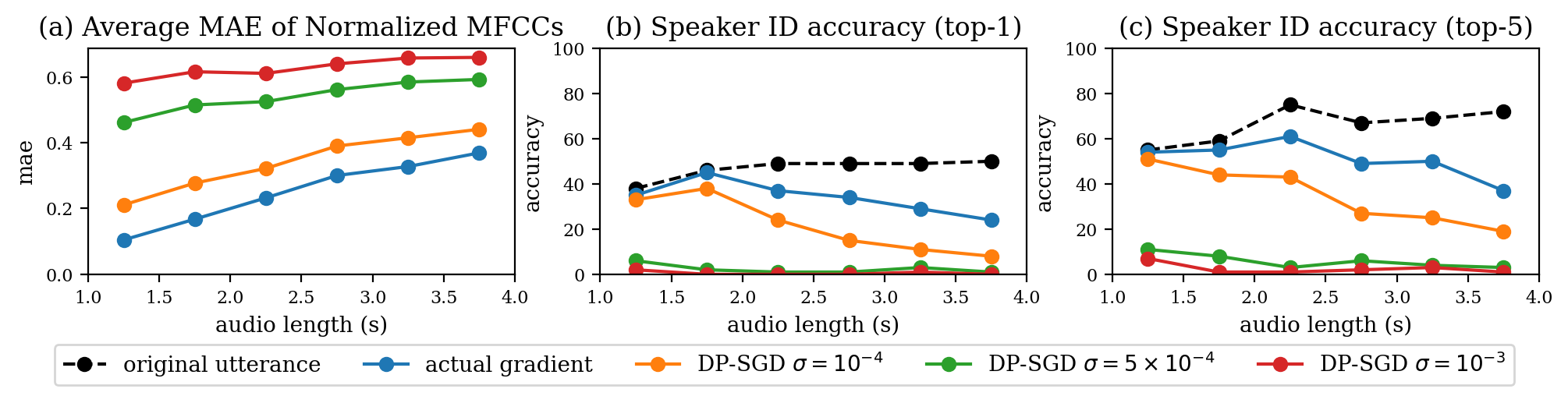}}
\caption{MAE and speaker identification accuracy on 600 utterances reconstructed using the actual gradient, and using DP-SGD at different noise levels for training. We also provide the respective accuracy for the original utterances. For short utterances, speaker identification on our reconstructions from the actual gradients almost matches that from the original utterances. For $\sigma=10^{-3}$, the top-1 accuracy reduces to $\sim$0\%.}
\label{fig:train-dp}
\end{center}
\end{figure*}

\subsection{Empirical Results}
\label{sec:rec_res}
Now, we present the results of using our method from Section~\ref{sec:method} to reveal speaker identity from 600 individual gradients, each gradient computed using a unique utterance from our sampled set.
In Figure~\ref{fig:train-dp}, we plot the results by audio length in intervals of 0.5s from 1-4s.
First, we show the average MAE of the reconstructed MFCCs, where we observe that the average MAE monotonously increases with the audio length.
Note that the dimensionality of the optimization problem increases linearly with the audio length.
Next, we plot the top-1 and top-5 accuracy of the speaker identity from the reconstructed MFCCs. 
Notice that even for the longest 3.5-4s utterances under consideration, the top-1 accuracy is 24\%, and the top-5 accuracy is 37\%. 
For comparison, we also plot the performance of our speaker identification model on the original utterances.
For shorter utterances, the top-1 accuracy from the reconstructed MFCCs is almost identical to that from the original utterances.

Table~\ref{tab:acc} shows the  overall values of the average MAE, Top-1 accuracy, Top-5 accuracy, and MRR of speaker identification results from the original and reconstructed speech features.
We see that while speaker identification from original utterances results in 42\% top-1 (57\% top-5) accuracy, the same from the reconstructed features is 34\% top-1 (51\% top-5), providing 81\% (89.5\%) relative performance.

\begin{table}[h]
  \caption{MAE, Top-1 Accuracy, Top-5 Accuracy, and MRR of speaker identification on 600 utterances. The top-1 (top-5) accuracy of speaker identification on reconstructed features is 81\% (89.5\%) relative to that on original utterances.}
  \label{tab:acc}
  \vskip 0.15in
  \begin{center}
\begin{small}
\begin{sc}
  \begin{tabular}{lcccc}
    \toprule
     & MAE & Top-1 & Top-5 & MRR                  \\
    \midrule
    Original & 0.00 & 42.0 & 57.0 & 0.554  \\
    Reconstructed     & 0.25 & 34.0 & 51.0 & 0.419 \\
    \bottomrule
  \end{tabular}
  \end{sc}
  \end{small}
  \end{center}
  \vskip -0.1in
\end{table}

%% file: sections/defense.tex
\subsection{Training with DP-SGD}
\label{sec:dp}
Now, we study the effect of training with the popular technique of Differentially Private Stochastic Gradient Descent (DP-SGD) on the success of our method.
At a high-level, each gradient gets \emph{clipped} to a fixed $L_2$-norm bound $C$, and zero-mean Gaussian noise of standard deviation $\sigma C$ is added to provide a (local) DP guarantee for each sample.
Due to space constraints, we defer the formal definition of Differential Privacy (DP)~\cite{DMNS, ODO}, and a pseudo-code of DP-SGD, to Appendix \ref{appendix:dpsgd}.

Using only $L_2$-clipping has been shown in prior works~\cite{carlini2019secret, thakkar2020understanding} to be effective in mitigating \emph{unintended memorization} in language models.
However, the optimization in our method (Equation~\ref{eqn:prob}) uses cosine distance as the loss, thus rendering only $L_2$-clipping ineffective.
Since using DP-SGD for training large models has been shown~\cite{abadi2016deep,mcmahan2017learning,thakkar2020understanding} to affect model utility, our first objective is to find the least $\sigma$ s.t. the top-1 accuracy of speaker identification is $\sim0$\%.
For our experiments, we set $C=100$\footnote{We observed gradients had norm at least 100, and thus chose $C=100$. Due to cosine loss used in our optimization, as long as a gradient gets clipped, the value of $C$ will not have any effect on the success of the method, or on the DP guarantee via DP-SGD.}, and provide the evaluation metrics for $\sigma \in \{10^{-4}, 5\times10^{-4}, 10^{-3}\}$ in Figure~\ref{fig:train-dp}.
We observe that $\sigma\geq 10^{-3}$ is effective in reducing the top-1 accuracy of speaker identification to $\sim0$\%. 

\bgroup
\def\arraystretch{1}
\setlength\tabcolsep{2.7pt}
\begin{table}[ht]
  \caption{MAE and Speaker identification from reconstruction when using DP-SGD at different noise levels. The noise needed ($10^{-3}$) to get $\sim$0\% top-1 accuracy almost doubles the final model's WER.}
  \label{tab:dp-train-res}
  \vskip 0.15in
  \centering
  \begin{center}
\begin{small}
\begin{sc}
  \begin{tabular}{lcccccc}
    \toprule
    $\sigma$ & MAE & Top-1 & Top-5 & MRR & \shortstack{WER\\(clean)} & \shortstack{WER\\(other)}                  \\
    \midrule
    Baseline     & 0.25 & 34.0 & 51.0 & 0.419 & 10.5 & 28.4 \\
    \midrule
    $10^{-4}$     & 0.34 & 21.5 & 34.8 & 0.284 & 14.9 & 37.6 \\
    $5\times 10^{-4}$ & 0.54 & 2.3 & 5.8 & 0.049 & 15.4 & 39.4 \\
    $10^{-3}$     & 0.63 & 0.5 & 1.7 & 0.021 & 19.6 & 45.3 \\
    \bottomrule
  \end{tabular}
  \end{sc}
  \end{small}
  \end{center}
  \vskip -0.1in
\end{table}
\egroup

We also provide the overall evaluation metrics with using DP-SGD in Table~\ref{tab:dp-train-res}.
We see that for $\sigma=10^{-3}$, the top-1 accuracy of speaker identification is 0.5\%.
Further, we  also provide the WER of models trained using DP-SGD~\cite{TFpriv} with a batch size of 16.
We see that the WER of models trained via DP-SGD, even for the smallest noise level, is significantly increased compared to our baseline training.
Note that for the levels of noise presented here, the bounds for (local/central) DP will be near-vacuous.
However, improving the privacy-utility trade-offs for DP-SGD is beyond the scope of this work.

\begin{figure*}[htb]
\centering
\includegraphics[width=0.49\textwidth]{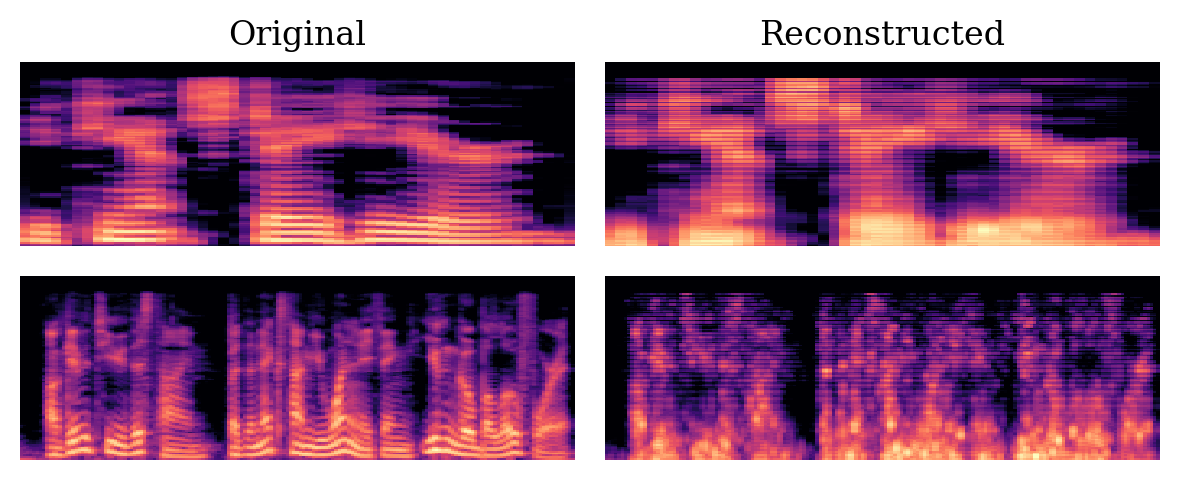}
\includegraphics[width=0.49\textwidth]{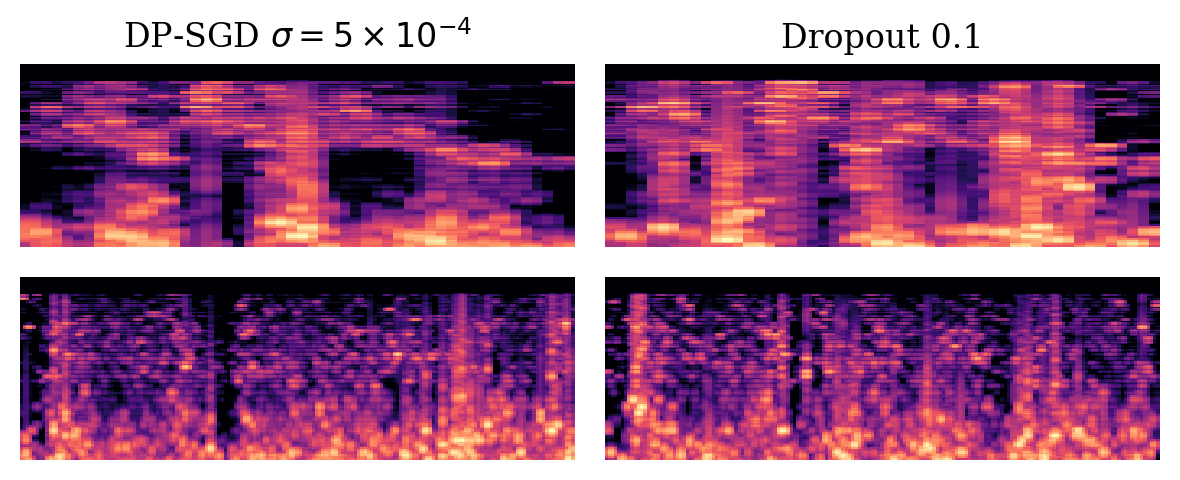}
\caption{Spectrograms obtained from the original and reconstructed MFCCs, as well as training with DP-SGD $\sigma=5\times 10^{-4}$, and dropout rate 0.1. 
The utterance in the first row is ``where is my husband'' (length: 1.4s, MAE of Reconstructed: 0.09, speaker identified correctly). The  utterance in the second row is ``i'll give it to you this time but the next time you want anything you can go below for it'' (length: 4.0s, MAE of Reconstructed: 0.31, speaker identified incorrectly). 
Even though the latter has a bad reconstruction quality, its spectrogram for the reconstructed features is visibly similar to that of the original. For reconstructions on training using DP-SGD, and Dropout, we see that reconstruction quality deteriorates.}
\label{fig:spectrogram}
\end{figure*}

\subsection{Training with Dropout}
\label{sec:dr}

Dropout \cite{srivastava2014dropout} has been adopted in training deep neural networks as an efficient way to prevent overfitting to the training data. 
The key idea of unit dropout is to randomly drop model units during training. 
While prior work~\cite{wei2020framework} has mentioned dropout in the context of information leakage from gradients, it does not provide any empirical evidence of the effect of training with dropout on such leakages.

The dropout mask is deducible from gradients if dropping a unit completely disables a part of the network (e.g. a feed-forward neural network), or dropout is applied directly on weights \cite{wan2013regularization}. When parameters are shared in the network, for e.g., a fully-connected layer operating frame-wise on a sequence of speech features, each part of the output typically uses an i.i.d. random dropout mask, making it difficult to infer dropout masks from a gradient.

\bgroup
\def\arraystretch{1}
\setlength\tabcolsep{4.5pt}
\begin{table}[h]
  \caption{MAE, Top-1, Top-5, and MRR of speaker identification when reconstructed from dropped-out gradients. Even a dropout rate of 0.1 efficiently prevents the leakage.}
  \label{tab:dropout-train-res}
  \vskip 0.15in
  \begin{center}
\begin{small}
\begin{sc}
  \begin{tabular}{lcccccc}
    \toprule
    $d$ & MAE & Top-1 & Top-5 & MRR & \shortstack{WER\\(clean)} & \shortstack{WER\\(other)}                  \\
    \midrule
    $0$     & 0.25 & 34.0 & 51.0 & 0.419 & 10.5 & 28.4 \\
    \midrule
    $0.1$     & 0.59 & 0.8 & 2.0 & 0.019 & 11.9 & 28.2 \\
    $0.2$     & 0.72 & 0.0 & 0.5 & 0.006 &  9.2 & 25.6 \\
    $0.3$     & 0.81 & 0.1 & 0.3 & 0.005 & 9.5 & 27.1 \\
    \bottomrule
  \end{tabular}
  \end{sc}
  \end{small}
  \end{center}
  \vskip -0.1in
\end{table}
\egroup

Table \ref{tab:dropout-train-res} shows reconstruction quality and training error rates for different dropout rates. 
Even for the lowest dropout rate of 0.1, we see that the top-1 accuracy of speaker identification is $\sim$0\%.
At the same time, we observe that for models trained with dropout, the WER is comparable (or sometimes even lower) than the baseline training.
We defer the plots of the results grouped by audio length to Appendix \ref{app:expt}.

\paragraph{Visualizing Reconstructed Features}
In Figure~\ref{fig:spectrogram}, we provide two examples of spectrograms from the reconstruction of a short and a long utterance. 
For the long utterance, even though MAE for the reconstruction is high and the speaker identification system fails to identify the speaker, the reconstructed audio pattern is visibly similar to the original audio pattern.
For comparison, we also provide  spectrograms from reconstructions of the same utterances from DP-SGD training ($\sigma=5\times 10^{-4}$), and a dropout rate of 0.1.

%% file: sections/multi.tex
\section{Additional Experiments}
\label{sec:multi}

The experiments in Section~\ref{sec:results} focused on revealing speaker identity using our method on a single gradient from a single utterance.
In distributed settings like FL, model training is performed under more complex settings.
In this section, we conduct experiments to evaluate the success of our method on two natural extensions of the setting in Section~\ref{sec:results}:
1) gradients from a batch of utterances are averaged before being shared, and 
2) multiple update steps are performed using a single utterance, and the final model update is shared.
We demonstrate that in both of the settings above, our method can reveal speaker identity with non-trivial accuracy.
Further, we show that using dropout for training reduces the limited success of the method in both the settings.
All the experiments in this section are conducted using the 200 utterances of audio length 1-2s (from the 600 sampled utterances for experiments in Section~\ref{sec:results}).

\subsection{Averaged Gradients from Batches}
\label{sec:avg}

In this section, we study the performance of our method for revealing speaker identities from an averaged gradient computed using a batch of utterances.
In the reconstruction phrase, our objective function \eqref{eqn:prob} does not change; however, we instead try to reconstruct $(x_1:x_N)$, where $x_i \in \mathbb{R}^{T_i\times d}$ for $i\in [B]$. Here, $B$ is the number of samples in the batch, and $T_i$ is the length of input $i \in [B]$. 
For computational efficiency, we only update a single sample per iteration of our optimization. 
We provide a pseudo-code for the variant of Algorithm~\ref{alg:dfgm} adapted to this setting, in Appendix \ref{appendix:multisample}.

We conduct our experiments for batch sizes in $\{2, 4, 8\}$. 
For each batch size, the 200 utterances are sorted by audio length, and grouped into batches. 
We provide the results in Table \ref{tab:multi-sample-acc}, comparing them with the results (batch size 1) on same 200 utterances in Section~\ref{sec:rec_res}.
We see that while speaker identification accuracy decreases with increasing batch sizes, the top-1 accuracy is still as high as 19\% for batch size 4.
An experiment on the effect of training with a dropout rate of 0.1 shows that reconstruction of batch size 2 from dropped-out gradients reduces the accuracy to 1\% top-1 (4\% top-5), compared to 2\% top-1 (4\% top-5) on the same set of utterances in Section~\ref{sec:dr}.

\begin{table}[ht]
  \caption{Reconstruction MAE, Top-1 and Top-5 Speaker identification accuracy from averaged gradients of a batch.
  Even with batch size 4, our method is successful with a top-1 accuracy of 19\%.}
  \label{tab:multi-sample-acc}
  \vskip 0.15in
  \begin{center}
\begin{small}
\begin{sc}
  \begin{tabular}{lccccc}
    \toprule
      & MAE & Top-1 & Top-5 & MRR                  \\
    \midrule
    Original     & 0.00 & 42.0 & 57.0 & 0.490 \\
    Batch size 1  & 0.14 & 40.0 & 55.0 & 0.470  \\
    \midrule
    Batch size 2      & 0.21 & 37.0 & 54.0 & 0.451  \\
    Batch size 4      & 0.37 & 19.0 & 31.0 & 0.249 \\
    Batch size 8      & 0.48 & 5.0 & 11.0 & 0.084 \\
    \bottomrule
  \end{tabular}
  \end{sc}
  \end{small}
  \end{center}
  \vskip -0.1in
\end{table}

\subsection{Multi-Step Updates from a Sample}

Now, we study the success of our method in revealing speaker identities from an update comprising of multiple update steps using a single utterance. 
We conduct our experiments for 2-step and 8-step updates with the learning rate set to $10^{-5}$.
For computational efficiency, we reduce the number of unit vectors sampled to 8 (as opposed to 128, in the experiments in Sections  \ref{sec:results} and \ref{sec:avg}) in each iteration of our zeroth-order optimization.

Table \ref{tab:multi-step-acc} shows the results of our experiment, comparing them with the same (1-step) from Section~\ref{sec:rec_res}. 
Since the optimization for multi-step reconstruction is different, the results are not directly comparable with those of single-step setting.
We see that even though the time/computation taken for reconstruction may increase with increasing number of steps, the success of our method in revealing speaker identity is still as high as 24\% top-1 accuracy for 8-step updates.
Using dropout in training is still effective:  a dropout rate of 0.1 reduces the accuracy to 2\% top-1 (3.5\% top-5).

\begin{table}[ht]
  \caption{Reconstruction MAE, Top-1 and Top-5 Speaker identification accuracy from multi-step updates from a single sample. We see that increasing the number of steps from 2 to 8 does not significantly affect the quality of the reconstruction}
  \label{tab:multi-step-acc}
  \vskip 0.15in
  \begin{center}
\begin{small}
\begin{sc}
  \begin{tabular}{lcccc}
    \toprule
     & MAE & Top-1 & Top-5 & MRR                  \\
    \midrule
    Original    & 0.00 & 42.0 & 57.0 & 0.490 \\
    1-step     & 0.14 & 40.0 & 55.0 & 0.470 \\
    \midrule
    2-step     & 0.33 & 26.5 & 39.5 & 0.333  \\
    8-step     & 0.33 & 24.5 & 39.0 & 0.321 \\
    
    \bottomrule
  \end{tabular}
  \end{sc}
  \end{small}
  \end{center}
  \vskip -0.1in
\end{table}

%% file: sections/prior.tex
\section{Related Work}
\label{sec:prior}
While we provide a background (in Section \ref{sec:background}) for  the DLG method \cite{zhu2019deep} and a comparison with our method (in Section~\ref{sec:comp}), there have been follow-up works~\cite{geiping2020inverting, wei2020framework,zhao2020idlg} showing high-fidelity image and label reconstruction from gradients under different settings.
Revealing information about training data from gradients has also been shown via membership and property leakage \cite{shokri2017membership,song2019auditing,melis2019exploiting}.
There is a growing line of works on revealing information from trained models.
For instance, \cite{fredrikson2015model} demonstrate vulnerabilities to model inversion attacks. Other works~\cite{carlini2019secret,thakkar2020understanding} show the amount of unintended memorization in trained models, along with studying the effect of  DP-SGD in mitigating such memorization.

For using standard training techniques to reduce information leakages from model training, while gradient compression and sparsification have been claimed \cite{zhu2019deep} to provide protection, it has been shown in \cite{wei2020framework} that reconstruction attacks can succeed with non-trivial accuracy in spite of using gradient compression.
There also exist works on designing strategies that require changes to the model inputs or architecture for protection, e.g., TextHide \cite{huang2020texthide}, and InstaHide \cite{huang2020instahide}.
For real-world deployments of distributed training, there also exist protocols like Secure Aggregation~\cite{secureagg} which make it difficult for any adversary to access raw individual gradients.

%% file: sections/acks.tex
\section*{Acknowledgements}
\label{sec:acks}
The authors would like to thank Nicholas Carlini, Andrew Hard, Ronny Huang, Khe Chai Sim, and our colleagues in Google Research for their helpful support of this work, and comments towards improving the paper.

%% file: sections/appendix/background.tex
\section{Background}

\subsection{DeepSpeech}

We hereby present details about the DeepSpeech \cite{hannun2014deep} model. The model consists of three feed-forward layers, followed by a single bi-directional LSTM layer, and two feed-forward layers to produce softmax probabilities for the CTC loss. The list of layers and number of parameters at each layer are shown in Table \ref{tab:deepspeech}. Note that we only use the last layer to match gradients, which has only $\sim0.1M$ parameters.

\bgroup
\def\arraystretch{1}
\setlength\tabcolsep{4.5pt}
\begin{table}[ht]
  \caption{Number of parameters at each layer of DeepSpeech. Note that we only match the last layer (Layer 6) during the reconstruction.}
  \label{tab:deepspeech}
  \vskip 0.15in
  \begin{center}
\begin{small}
\begin{sc}
  \begin{tabular}{llc}
    \toprule
    Layer & Type & No. parameters \\
    \midrule
    1 & feed-forward & 1.0m \\
    2 & feed-forward     & 4.2m \\
    3 & feed-forward     & 4.2m \\
    4 & bi-directional lstm     & 33.6m \\
    5 & feed-forward     & 4.2m \\
    6 & feed-forward     & 0.1m \\
    \midrule
    total & & 47.3m \\
    \bottomrule
  \end{tabular}
  \end{sc}
  \end{small}
  \end{center}
  \vskip -0.1in
\end{table}
\egroup

\subsection{Deep Speaker}

The Deep Speaker \cite{li2017deep} model adopts a deep residual CNN (ResCNN) architecture to extract the acoustic features from utterances. These per-frame features are averaged to produce utterance-level speaker embeddings. The ResCNN consists of four stacked residual blocks (ResBlocks) with a stride 2. The numbers of CNN filters are 64, 128, 256, 512, respectively. The total number of parameters is 24M.

Deep Speaker is trained with Triplet Loss, which takes three samples as input, an anchor $a$, a positive sample $p$ (from the same speaker), and a negative sample $n$ (from another speaker). The loss function of $N$ samplings is defined as

\[L=\sum_{i=0}^N\max(0, s_i^{an}-s_i^{ap}+\alpha)\]

where $s_i^{an}$ is the cosine similarity between the anchor $a_i$ and the negative sample $n_i$, $s_i^{ap}$ is the cosine similarity between the anchor $a_i$ and the positive sample $p_i$, from the $i$-th sampling. $\alpha$ is the minimum margin between these cosine similarities, which is set to 0.1.

\subsection{DP-SGD}
\label{appendix:dpsgd}

For completeness, we start by providing a definition of the notion of Differential Privacy~\cite{DMNS, ODO}. We will refer to a pair of datasets $X,X'$ as neighbors if $X'$ can be obtained by the addition or removal of one sample from $X$.

\begin{definition}[Differential privacy~\cite{DMNS, ODO}] A randomized  algorithm $\mathcal{A}$ is $(\epsilon,\delta)$-differentially private if, for any pair of neighboring datasets $X$ and $X'$, and for all events $\mathcal{S}$ in the output range of $\mathcal{A}$, we have 
$$\Pr[\mathcal{A}(X)\in \mathcal{S}] \leq e^{\epsilon} \cdot \Pr[\mathcal{A}(X')\in \mathcal{S}] +\delta$$
where the probability is taken over the random coins of $\mathcal{A}$. 
\label{def:diiffP}
\end{definition}

Now, we provide a pseudo-code for DP-SGD~\cite{abadi2016deep}.

\begin{algorithm}[ht]
	\begin{algorithmic}[1]
		\REQUIRE Dataset $X$ of size $n$, Loss function $\mathcal{L}(\theta)$, Parameters: Mini-batch size $B$, Learning rate $\eta$, Clip norm bound $C$, Per-sample noise scale $\sigma$, Total number of iterations $T$
		
		\STATE Initialize model with $\theta_0$\ randomly\;
		\FOR{$t\in [1,T]$}
		\STATE Sample a random minibatch $B_t \subseteq X$, by independently including each element of $X$ with probability $B/n$\; 
		\FOR{$x_i \in B_t$}
		\STATE Compute gradient  $g_t(x_i) = \nabla_\theta\mathcal{L}(\theta_t,x_i)$\;
		\STATE Clip each gradient in $\ell_2$ norm to $C$, i.e.,  $\bar{g}_t(x_i) = g_t(x_i)/\max(1, \frac{\|g_t(x_i)\|_2}{C}) $\;
		\STATE Add noise $\tilde{g}_t(x_i) =  \bar{g}_t(x_i) + \mathcal{N}(0,\sigma^2C^2I)$\;
		\ENDFOR
		\STATE Compute average noised gradient $\hat{g}_t =  \frac{1}{|L|} (\sum_i \tilde{g}_t(x_i)$
		\STATE Update model $\theta_t = \theta_{t-1} - \eta \cdot \hat{g}_t$ \;
		\ENDFOR
		\STATE Compute privacy cost using Moments Accountant.
		\caption{Differentially Private SGD}
		\label{algo:dpopt}
	\end{algorithmic}
\end{algorithm}

%% file: sections/appendix/expt.tex
\section{Additional Experiments, and Omitted Details}
\label{app:expt}

\subsection{Reconstruction without Assumptions}
\label{app:assump}

We set up two experiments to explore the necessity of the two assumptions of known input length and transcript for our reconstruction method.

\paragraph*{Reconstruction without Knowledge of Input Length}
\label{app:inp}

The first assumption for our reconstruction method (Section~\ref{sec:dfgm}) that the length of input speech features is known. This is required to set up the search space for the optimization problem \eqref{eqn:prob}. Without the exact input length, we show that reconstruction is still possible. In the experiments below, 20 random utterances are chosen from the 1-2s bucket whose speaker are correctly identified in top-5 in section~\ref{sec:rec_res}. The average length of these utterances is 74.35 frames ($\sim$ 1.5s). Table \ref{tab:no-input-len} shows reconstruction results when estimated lengths differ by $\pm 1, \pm 5$, and $\pm 10$ compared to original lengths and are double / half of the original lengths. It can be seen that the speaker identity can still be revealed even with a good estimate of the input length. For the same amount of absolute deviation in the estimation (e.g., $+5$ and $-5$), we see that the higher estimation provides better results.

\bgroup
\def\arraystretch{1}
\setlength\tabcolsep{4.5pt}
\begin{table}[ht]
  \caption{Loss (gradients' distance) and speaker identification results with different input lengths on 20 random short utterances correctly identified top-5 speaker in section~\ref{sec:rec_res}. We see that our method succeeds even with good estimates of the input length.}
  \label{tab:no-input-len}
  \vskip 0.15in
  \begin{center}
\begin{small}
\begin{sc}
  \begin{tabular}{lcccc}
    \toprule
    Length & Loss ($\times 10^{-3}$) & Top-1 & Top-5 & MRR \\
    \midrule
    Original & 0.04 & 90 & 100 & 0.748 \\
    \midrule
    $+1$ & 0.06 & 60 & 90 & 0.706 \\
    $-1$ & 0.05 & 55 & 95 & 0.714 \\
    $+5$ & 0.20 & 50 & 80 & 0.632 \\
    $-5$ & 0.31 & 45 & 70 & 0.580 \\
    $+10$ & 0.44 & 35 & 55 & 0.442 \\
    $-10$ & 1.43 & 20 & 40 & 0.301 \\
    \midrule
    $\times 2$ & 2.53 & 0 & 10 & 0.048 \\
    $/ 2$ &  145.15   & 0 & 0 & 0.003 \\
    \bottomrule
  \end{tabular}
  \end{sc}
  \end{small}
  \end{center}
  \vskip -0.1in
\end{table}
\egroup

\paragraph*{Reconstruction without Contents of the Transcript}

Next, we conduct experiments with our method having knowledge of only the length of the transcript, not its contents. For each utterance in the set of 20 utterances from section~\ref{app:inp}, we generate 4 random transcripts and use them to reconstruct speech features. It can be seen from Table \ref{tab:no-transcript} that reconstruction is constantly of a poor quality (high loss) with a random transcript, suggesting that the knowledge about the transcript is important. The bad quality of reconstructed features from an incorrect transcript also suggests that if the attacker has a list of candidates for the transcript (e.g., common phrases, song names, etc.) including the original one, a brute-force approach to pick the one with the lowest loss can reveal the actual transcript with high confidence.

\bgroup
\def\arraystretch{1}
\setlength\tabcolsep{3.0pt}
\begin{table}[ht]
  \caption{Loss and speaker identification results when reconstructing from random transcripts of the same length as the original. Reconstruction is constantly of a poor quality with a random transcript.}
  \label{tab:no-transcript}
  \vskip 0.15in
  \begin{center}
\begin{small}
\begin{sc}
  \begin{tabular}{lccccc}
    \toprule
    Transcript & Loss ($\times 10^{-3}$) & MAE & Top-1 & Top-5 & MRR \\
    \midrule
    Original & 0.04 & 0.12 & 90 & 100 \\
    \midrule
    Random 1 & 79.5 & 0.78 &  0 & 0 & 0.010 \\
    Random 2 & 135.5 & 0.74 & 0 & 0 & 0.013  \\
    Random 3 & 108.7 & 0.77 & 0 & 0 & 0.006 \\
    Random 4 & 101.5 & 0.78 &  0 & 0 & 0.015 \\
    \bottomrule
  \end{tabular}
  \end{sc}
  \end{small}
  \end{center}
  \vskip -0.1in
\end{table}
\egroup

\subsection{Reconstruction from Dropped-Out Gradients}

\begin{figure*}[ht]
\begin{center}
\centerline{\includegraphics[width=.9\textwidth]{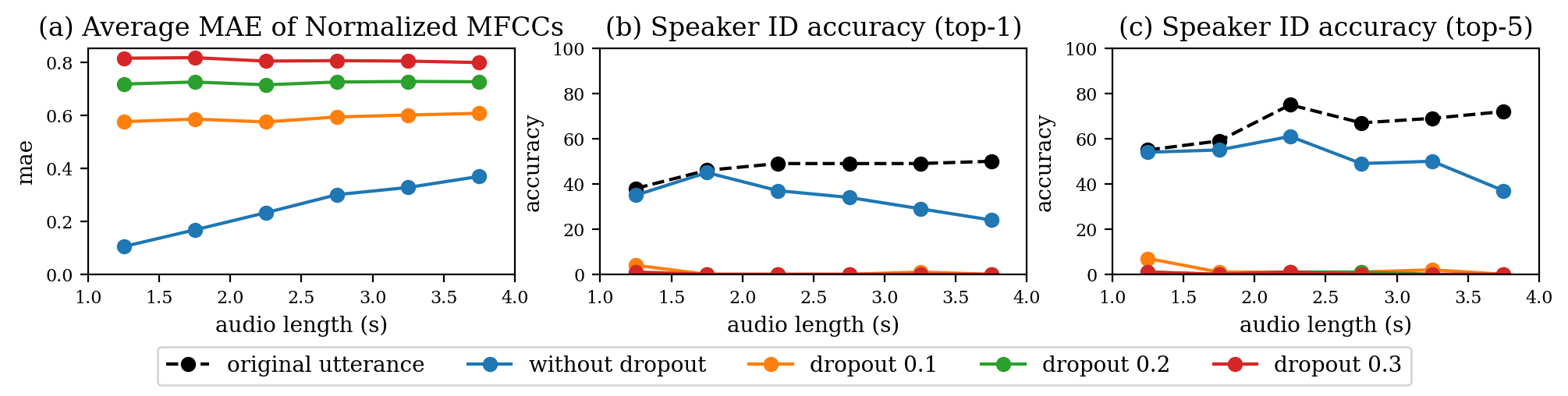}}
\caption{ MAE and speaker identification accuracy when reconstructed from dropped-out gradients. Even a dropout rate of 0.1 efficiently reduces the speaker identification accuracy to 0.}
\label{fig:graph-acc-by-length-dropout}
\end{center}
\end{figure*}

Figure \ref{fig:graph-acc-by-length-dropout} show results grouped by audio length of the experiment in Section \ref{sec:dr}. Even a dropout rate of 0.1 efficiently eliminates the risk of speaker identity leakage. 

\begin{figure*}[ht]
\begin{center}
\centerline{\includegraphics[width=.9\textwidth]{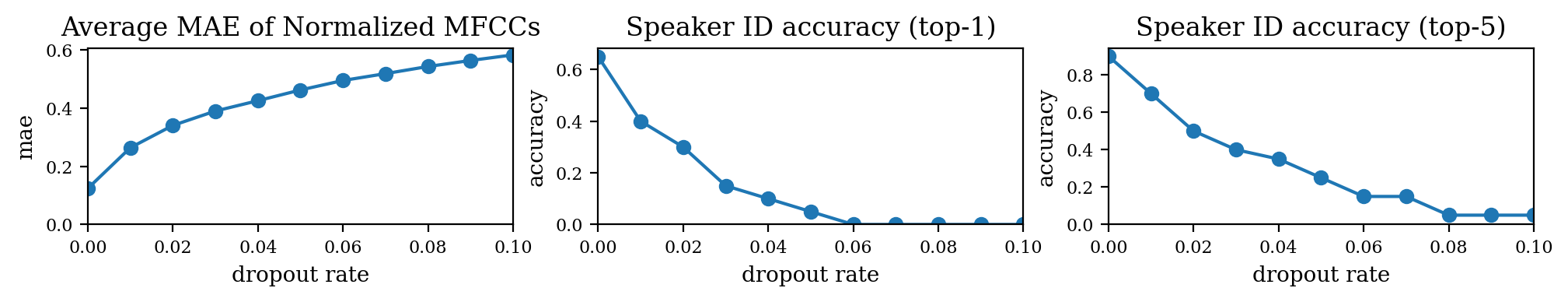}}
\caption{MAE and speaker identification accuracy when dropout rate changes from 0.00 to 0.10. Each point is averaged from 20 short utterances. Speaker identification accuracy drops sharply when increasing the dropout rate.}
\label{fig:graph-acc-by-length-dropout-20}
\end{center}
\end{figure*}

We also try varying the dropout rate and performing reconstruction on a small population of 20 utterances (first 20 utterances when sorted by lengths). The results are presented in Figure \ref{fig:graph-acc-by-length-dropout-20}. The speaker identification accuracy drops sharply when increasing the dropout rate.

%% file: sections/appendix/app_proposed.tex
\section{Algorithms}
We present an adapted version of \shortname{} for reconstructing from averaged gradients and multi-step updates.

\subsection{\shortname{} on Averaged Gradients from Batches}
\label{appendix:multisample}

In Algorithm \ref{alg:dfgm}, a dummy input is randomly initialized at the beginning and given to the model to compute the loss and gradients at every iteration of the optimization process. When reconstructing a batch from averaged gradients, a dummy batch needs to be optimized. To save computation time, we only update a single sample at each iteration, reusing the loss and gradients of other samples in the batch to obtain the overall loss and gradients. A variant of Algorithm \ref{alg:dfgm} adapted for this setting is presented as Algorithm \ref{alg:multisample}

\begin{algorithm}[ht]
   \caption{\shortname{} on averaged gradients from batches}
   \label{alg:multisample}
\begin{algorithmic}
   \STATE {\bfseries Input:} Gradients to match $\nabla\mathcal{L}(\hat{x}, \hat{y})$, gradients distance function $\mathcal{D}(x,\nabla\mathcal{L}(\hat{x},\hat{y}))$, learning rate $\alpha$, transcript $\hat{y}$, length of speech features $\{T_i\}_{i=1}^{B}$. 
   Parameters: number of samplings $K$, number of iterations $N$, batch size $B$
   \STATE Initialize $\{x_i\}_{i=1}^B$, $x_i\in\mathbb{R}^{T_i\times d}$.
   \STATE $\mathcal{L}_i\leftarrow\mathcal{L}(x_i, \hat{y}_i)$ for $i\in[B]$ 
   \STATE $\nabla \mathcal{L}_i\leftarrow\nabla\mathcal{L}(x_i, \hat{y}_i)$ for $i\in[B]$
   \FOR{$n=1$ {\bfseries to} $N$}
   \STATE $V\leftarrow\varnothing$
   \STATE Sample $b\in[B]$
   \STATE Sample $K$ column unit vectors $v_1,...,v_K\in\mathbb{R}^{T_i\times d}$
   \FOR{$k=1$ {\bfseries to} $K$}
        \STATE $x'\leftarrow\{x_1,...,x_b+\alpha v_k,..., x_B\}$
        \STATE $\mathcal{L}(x',\hat{y})\leftarrow\frac{1}{N}\left(\sum_{i\ne b}\mathcal{L}_{i} + \mathcal{L}(x_b+v_k,\hat{y_b})\right)$
        \STATE $\nabla\mathcal{L}(x',\hat{y})\leftarrow\frac{1}{N}\left(\sum_{i\ne b}\nabla\mathcal{L}_i + \nabla \mathcal{L}(x_b+v_k, \hat{y_b})\right)$
        \IF{$\mathcal{D}(x',\mathcal{L}(\hat{x},\hat{y}))<\mathcal{D}(x,\nabla\mathcal{L}(\hat{x}, \hat{y}))$}
            \STATE Add $v_i$ to $V$
        \ENDIF
   \ENDFOR
   \STATE $x_b\leftarrow x_b+\alpha \sum_{v\in V}v$
   \STATE $\mathcal{L}_b\leftarrow\mathcal{L}(x_b,\hat{y_b})$
   \STATE $\nabla\mathcal{L}_b\leftarrow\nabla\mathcal{L}(x_b,\hat{y_b})$
   \ENDFOR
\end{algorithmic}
\end{algorithm}

\subsection{\shortname{} on Multi-Step Updates from a Sample}

A challenge when applying Algorithm $\ref{alg:dfgm}$ to this setting is the change in model parameters after each local step. Therefore, model updates of sampled unit vectors cannot be computed in batch, but need to be computed separately. The model also needs to be reset to its original parameters before each computation. Algorithm $\ref{alg:multistep}$ provides a modified version of Algorithm $\ref{alg:dfgm}$ to reconstruct an input from multi-step updates. For efficiency, if $K$ vectors are sampled at each iteration, $K$ separate versions of the model are stored in the computation graph and processed in parallel.

\label{appendix:multistep}
\begin{algorithm}[ht]
   \caption{\shortname{} on multi-step updates from a sample}
   \label{alg:multistep}
\begin{algorithmic}
   \STATE {\bfseries Input:} Parameter changes to match $\Delta\theta$, gradients distance function $\mathcal{D}(x,\Delta\theta)$, learning rate $\alpha$, transcript $\hat{y}$ and length of speech features $T$. Parameters: number of samplings $K$, number of iterations $N$, number of steps $C$, local learning rate $\gamma$.
   \STATE Initialize $x\in\mathbb{R}^{T\times d}$.
   \FOR{$n=1$ {\bfseries to $N$}}
   \STATE $V\leftarrow\varnothing$
   \STATE Sample $K$ unit vectors $v_1,...,v_k\in\mathbb{R}^{T\times d}$
   \FOR{$k=1$ {\bfseries to} $K$}
        \STATE Reset $\theta_0$
        \FOR{$c=1$ {\bfseries to} $C$}
        \STATE
        $\theta_c\leftarrow\theta_{c-1}-\gamma\nabla\mathcal{L}_{\theta_{c-1}}(x+\alpha v_k, \hat{y})$
        \STATE Compute $\mathcal{L}_{\theta_c}(x+\alpha v_k,\hat{y})$ and  $\nabla\mathcal{L}_{\theta_c}(x+\alpha v_k,\hat{y})$
        \ENDFOR
        \IF{$\mathcal{D}(x+\alpha v_k,\Delta\theta)<\mathcal{D}(x,\Delta\theta)$}
            \STATE Add $v_k$ to $V$
        \ENDIF
   \ENDFOR
   \STATE $x\leftarrow x+\alpha \sum_{v\in V}v$
   \ENDFOR
\end{algorithmic}
\end{algorithm}

%% file: sections/appendix/visualize.tex
\section{Additional Visualizations}

\begin{figure*}[htb]
\begin{center}
\centerline{\includegraphics[width=\textwidth]{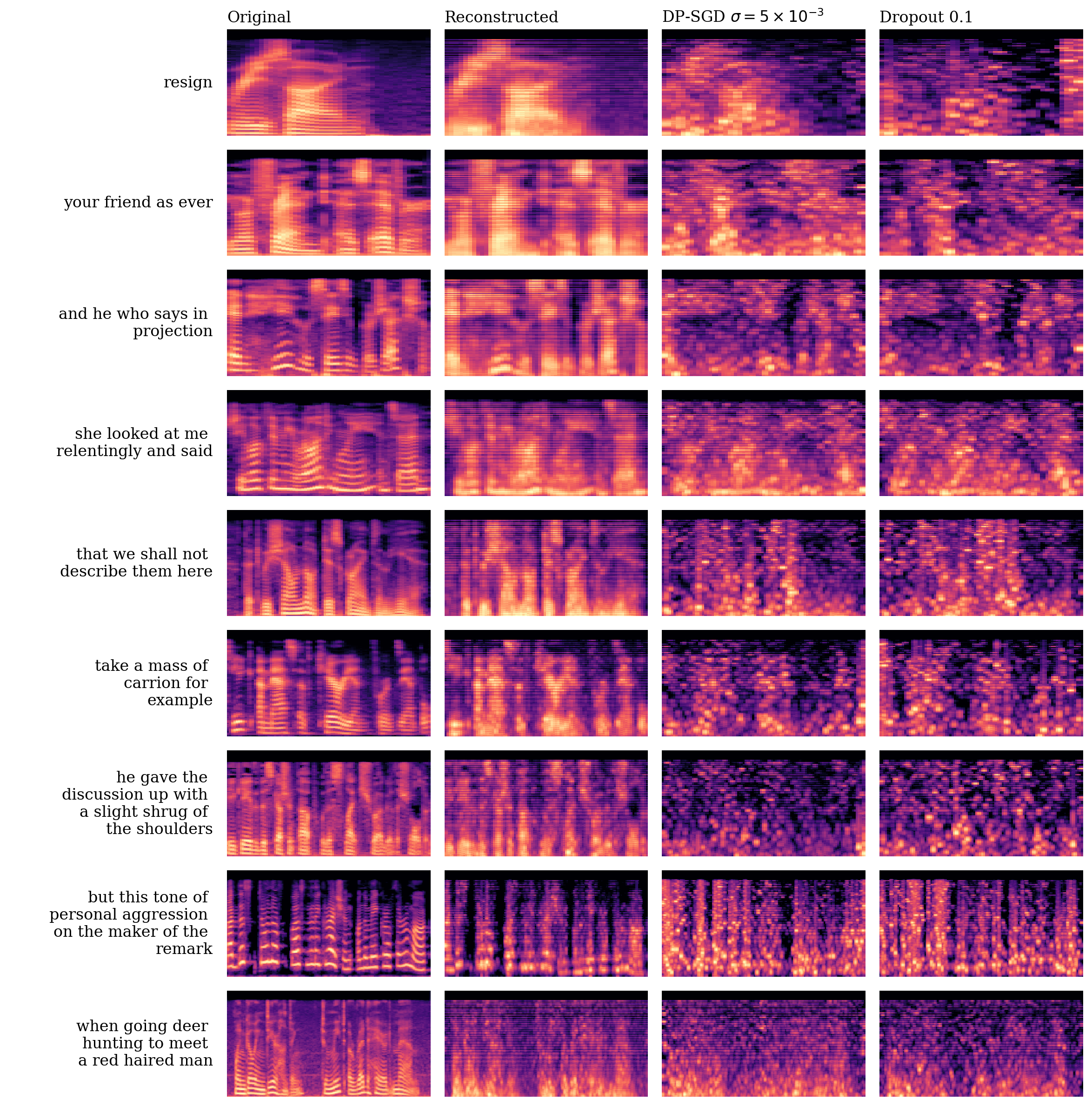}}
\caption{More spectrograms from reconstructions in Section \ref{sec:rec_res}. From left to right: Original, Reconstructed, DP-SGD with $\sigma=5\times 10^{-4}$, Dropout 0.1}
\label{fig:more-spec}
\end{center}
\end{figure*}

Figure \ref{fig:more-spec} shows the spectrogram of some utterances reconstructed in Section \ref{sec:rec_res}, along with results when reconstructing from a gradient with DP-SGD and Dropout.

\begin{figure*}[htb]
\begin{center}
\centerline{\includegraphics[width=\textwidth]{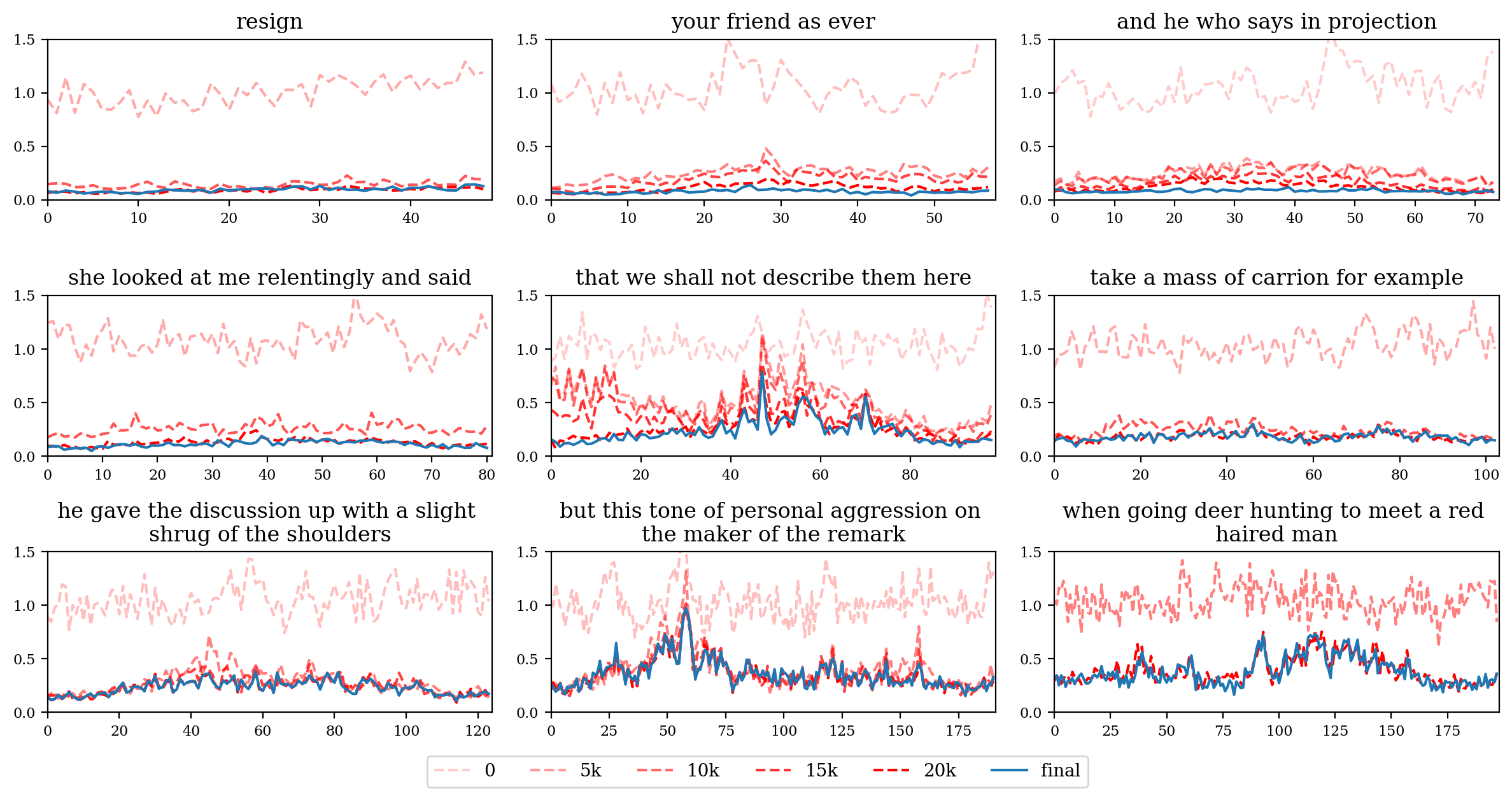}}
\caption{Per-frame MAEs of reconstructed MFCCs at different stages of the optimization. x-axis is the frame number and y-axis is the per-frame MAE. Darker lines mean more iterations have been run. The most blurred red line is the initial state and the blue line is the final result. For long utterances, frames at two ends are reconstructed better than those in the middle}
\label{fig:train-log}
\end{center}
\end{figure*}

We also plot per-frame MAEs at different stages in the optimization process in Figure \ref{fig:train-log}.
In long utterances, reconstructions usually have bad quality with frames in the middle being poorly reconstructed. This suggests that the error from earlier frames may have affected reconstruction in the middle part, due to sequential dependencies modeled in the LSTM.

